\documentclass[11pt,a4paper]{article}
\usepackage[hyperref]{emnlp-ijcnlp-2019}
\usepackage{times}
\usepackage{latexsym}
\usepackage{graphicx}
\usepackage{dblfloatfix}
\usepackage{amssymb}
\usepackage{multirow}
\usepackage{amsmath}
\usepackage{tabulary}
\usepackage{booktabs}

\usepackage{url}

\aclfinalcopy % Uncomment this line for the final submission

\title{TIGEr: Text-to-Image Grounding for Image Caption Evaluation}

\author{Ming Jiang$^{1}$, Qiuyuan Huang$^{3}$, Lei Zhang$^{3}$, Xin Wang$^{2}$ \\ \textbf{Pengchuan Zhang$^{3}$, Zhe Gan$^{3}$, Jana Diesner$^{1}$, Jianfeng Gao$^{3}$ }\\
  $^{1}$University of Illinois at Urbana-Champaign, $^{2}$University of California, Santa Barbara \\
  $^{3}$Microsoft Research, Redmond\\
  {\tt \{mjiang17,jdiesner\}@illinois.edu, xwang@cs.ucsb.edu} \\
  {\tt \{qihua,leizhang,penzhan,zhe.gan,jfgao\}@microsoft.com}}

\date{}

\begin{document}
\maketitle
\begin{abstract}
  This paper presents a new metric called TIGEr for the automatic evaluation of image captioning systems. Popular metrics, such as BLEU and CIDEr, are based solely on text matching between reference captions and machine-generated captions, potentially leading to biased evaluations because references may not fully cover the image content and natural language is inherently ambiguous. Building upon a machine-learned text-image grounding model, TIGEr allows to evaluate caption quality not only based on how well a caption represents image content, but also on how well machine-generated captions match human-generated captions. Our empirical tests show that TIGEr has a higher consistency with human judgments than alternative existing metrics. We also comprehensively assess the metric's effectiveness in caption evaluation by measuring the correlation between human judgments and metric scores.
\end{abstract}

\begin{figure}[th]
    %\centering
    \includegraphics[width = 0.48\textwidth]{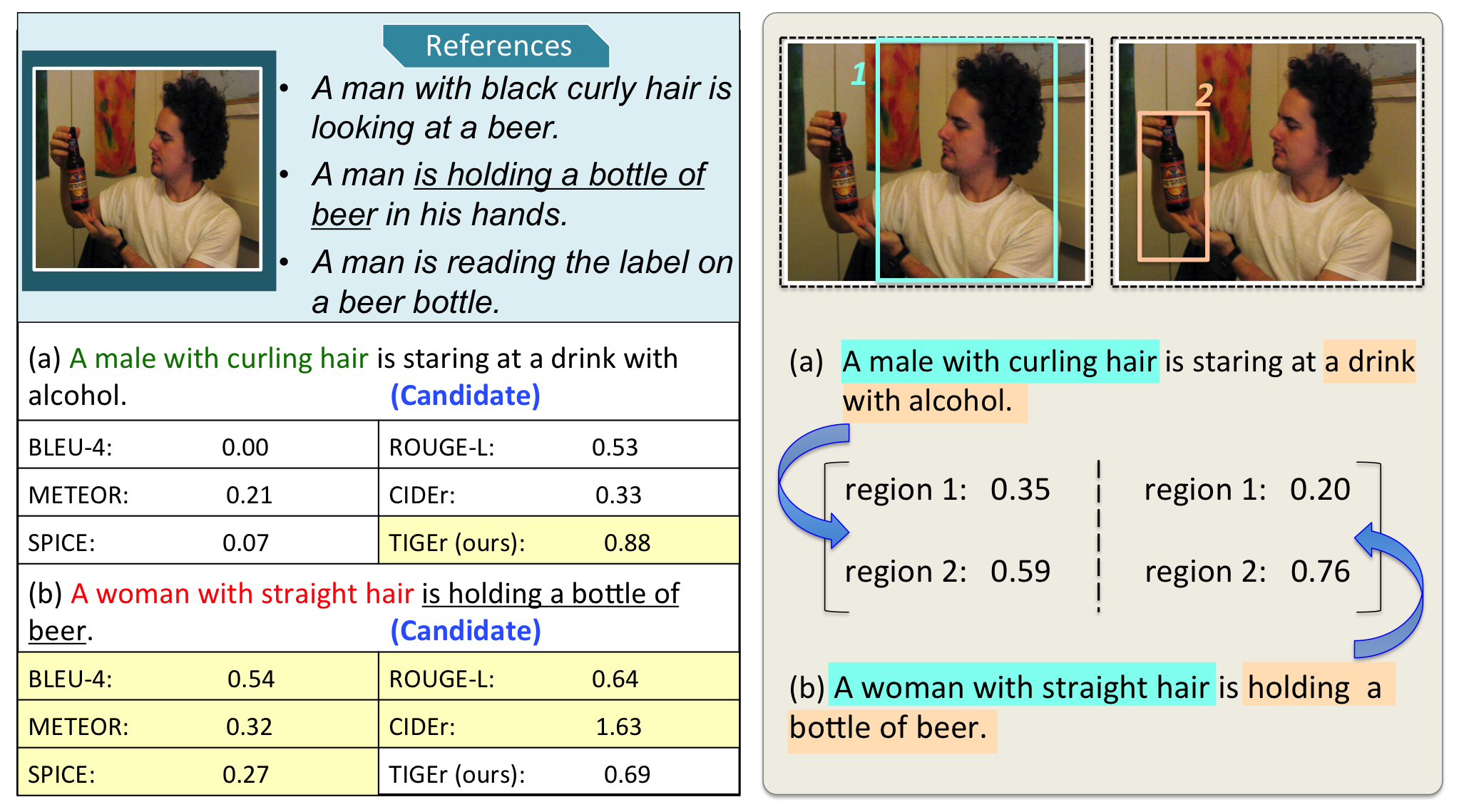}
    %\includegraphics[width = \textwidth]{Intro_Demo.pdf}
    %\vspace{-3ex}
    \vspace{-4ex}
    \caption{An example of caption evaluation challenge. Given two candidate captions: (a) correctly describes image content, but most words in this caption are not contained in references, while (b) overlaps with references (``is holding a bottle of beer"), but involves wrong information (``a woman with straight hair"). Prior metrics based only on text-level comparisons fail to assess the quality of candidate captions, while our metric (TIGEr) can detect this inconsistency by matching a caption with image content. To explain the TIGEr result, we show the grounding weights of two illustrative image regions with each candidate. The proper text information matched with each region is highlighted by colors.}
    \label{fig:demo}
    \vspace{-3ex}
\end{figure}

\section{Introduction}
%The joint study of language-vision  such as 
Image captioning is a research topic at the nexus of natural language processing and computer vision, and has a wide range of practical applications \cite{flickr8k, fang2015captions}, such as image retrieval and human-machine interaction. Given an image as input, the task of image captioning is to generate a text that describes the image content. Overall, prior studies of this task have been focusing on two aspects: 1) building large benchmark datasets \cite{MSCOCO, flickr8k}, and 2) developing effective caption generation algorithms \cite{captiongen2, captionsurvey, captiongen1}. Remarkable contributions that have been made, but assessing the quality of the generated captions is still an insufficiently addressed issue. 
Since numerous captions with varying quality can be produced by machines, it is important to propose an automatic evaluation metric that is highly consistent with human judges, easy to implement and interpret.
%we need to be able to automatically and hence efficiently evaluate if these auto-generated captions are consistent with human judgment, and easy to interpret.
%highly accurate, consistent with human judges, easy to implement, and interpretable.

%what are common properties of current techniques and are there any shortcomings of these properties?
Prior evaluation metrics typically considered several aspects, including content relevance, information correctness, grammaticality, and if expressions are human-like \cite{flickr8k,captionsurvey}. 
Rule-based metrics inspired by linguistic features such as n-gram overlapping are commonly used to evaluate machine-generated captions \cite{captiongen18, captiongen2, qiuyuan_hrl}. However, these metrics primarily evaluate a candidate caption based on references without taking image content into account, and the possible information loss caused by references may bring biases to the evaluation process. Moreover, the ambiguity inherent to natural language presents a challenge for rule-based metrics that are based on text overlapping. As shown in Figure~\ref{fig:demo}, although Caption (a) describes the image properly, most words in this sentence are different from the references. In contrast, Caption (b) contains wrong information (``a woman with straight hair"), but overlaps with human-written references (``is holding a bottle of beer"). Consequently, all existing metrics improperly assign a higher score to Caption (b).

In order to address the aforementioned challenges, we propose a novel Text-to-Image Grounding based metric for image caption Evaluation (TIGEr), which considers both image content and human-generated references for evaluation. First, TIGEr grounds the content of texts (both reference and generated captions) in a set of image regions by using a pre-trained image-text grounding model. Based on the grounding outputs, TIGEr calculates a score by comparing both the relevance ranking and distribution of grounding weights among image regions between the references and the generated caption. Instead of evaluating image captions by exact n-gram matching, TIGEr compares the candidate caption with references by mapping them into vectors in a common semantic space.

As a result, TIGEr is able to detect the paraphrases based on the semantic meaning of caption sentences. A systematic comparison of TIGEr and other commonly used metrics shows that TIGEr improves the evaluation performance across multiple datasets and demonstrates a higher consistency with human judgments.
For example, Figure~\ref{fig:demo} presents a case where only TIGEr is able to assign a higher score to the caption candidate that is more consistent with human judgments. 
Our main contributions include: 

\begin{itemize}
    \item We propose a novel automatic evaluation metric, TIGEr\footnote{Code is released at \url{https://github.com/SeleenaJM/CapEval}.}, to assess the quality of image captions by grounding text captions in image content.
    \item By performing an empirical study, we show that TIGEr outperforms other commonly used metrics, and demonstrates a higher and more stable consistency with human judgments.
    \item We conduct an in-depth analysis of the assessment of metric effectiveness, which deepens our understanding of the characteristics of different metrics. 
    %and provide a more comprehensive evaluation by revising commonly used correlation measurements. 
\end{itemize}

\section{Related Work}
% what is image caption evaluation task? How to define this task?
\paragraph{Caption Evaluation}
The goal of image caption evaluation is to measure the quality of a generated caption given an image and human-written reference captions \cite{captionsurvey}. 

%where the judging criteria include content relevance, information correctness and language expression in terms of grammar and fluency. 
% Existing evaluation methods (Human, rule-based metric, learning-based metric) What purpose? How to measure? Pros and Cons?
In general, prior solutions to this task can be divided into three groups. First, human evaluation is typically conducted by employing human annotators to assess captions (e.g., via Amazon’s Mechanical Turk) \cite{captionsurvey, composite, vstory}. %directly assesses captions by employing human annotators \cite{captionsurvey, composite}. Usually, this group of evaluation is conducted based on online surveys (e.g., Amazon’s Mechanical Turk) \cite{vstory, composite}. 

Second, automatic rule-based evaluation measures assess the similarity between references and generated captions. Many metrics in this group were extended from other related tasks \cite{bleu, meteor, rouge}. BLEU \cite{bleu} and METEOR \cite{meteor} were initially developed to evaluate machine translation outcomes based on n-gram precision and recall. ROUGE \cite{rouge} was originally used in text summarization, which measures the overlap of n-grams using recall. Recently, two metrics were specifically built for visual captioning: 1) CIDEr \cite{cider} measures n-gram similarity based on TF-IDF; and 2) SPICE \cite{spice} quantifies graph similarity based on the scene graphs built from captions. Overall, these metrics focus on text-level comparisons, assuming that the information contained in human-written references can well represent the image content. Differing from prior work, we argue in this study that references may not fully cover the image content because both references and captions are incomplete and selective translations of image contents made by human judges or automated systems. The ground truth can only be fully revealed by taking the images themselves into account for evaluation. 

Finally, machine-learned metrics use a trained model to predict the likelihood of a testing caption as a human-generated description \cite{learningmetric1, learningmetric2}. Prior studies have applied learning-based evaluation for related text generation tasks, e.g., machine translation \cite{machine_learn1, machine_learn2, machine_learn3}. 
% this type of study has just been extended to the task of image caption evaluation. 
Most recently, \citet{learningmetric1} trained a hybrid neural network model for caption evaluation based on image and text features. This work mainly focuses on the generation of adversarial data used for model training. Despite improving the consistency with human decisions, this approach may involve high computational cost and lead to overfitting \cite{gao2019neural}. Besides, the interpretability of the evaluation is limited due to the black-box nature of end-to-end learning models. An even more serious problem of using machine-learned metrics is the so-called ``gaming of the metric'': if a generation system is optimized directly for a learnable metric, then the system's performance is likely to be over-estimated. See a detailed discussion in \citet{gao2019neural}.  

Though caption evaluation is similar to traditional text-to-text generation evaluation in terms of testing targets and evaluation principles, this task additionally need to synthesize image and reference contents as ground truth information for further evaluation, which is hard to achieve by using a rule-based method with a strong human prior. In this study, we propose a new metric that combines the strengths of learning-based and rule-based approaches in terms of: 1) utilizing a pre-trained grounding model\footnote{We find in our experiments that we can simply use the pre-trained model as is to evaluate systems developed on different datasets, and the pre-trained model is very efficient to run. Thus, the cost of computing TIGEr is not much higher than that of computing other traditional metrics.}
to combine the image and text information;
%without adaption, fine-tuning and the dependency of datasets to combine image information with texts based on a manageable cost; %, and thus avoids the computational cost; 
and 2) scoring captions by defining principle rules based on the grounding outputs to make the evaluation interpretable.
%we then define principled rules to calculate a score based on the learned grounding outputs, so the caption evaluation process can be explained.  

% Related algorithms proposed for image-text matching?
\paragraph{Image-Text Matching}
The task of image-text matching can be defined as the measurement of semantic similarity between visual data and text data. The main strategy of solutions designed for this task is to map image data and text data into a common semantic vector space \cite{fang2015captions,imgtxt, scan}. Prior studies usually consider this issue from two perspectives:  
From the perspective of data encoding, prior work either regards an image/text as a whole or applies bottom-up attention to encode image regions and words/phrases \cite{imgtxt_encode1, imgtxt_encode2}. 
From the perspective of algorithm development, prior work has often applied a deep learning framework to build a joint embedding space for images and texts, and some of these studies train a model with ranking loss \cite{fang2015captions,scan, imgtxt_rank} and some use classification loss (e.g., softmax function) \cite{imgtxt_classification1, imgtxt_classification2}. In this work, we take advantage of a state-of-the-art model for image-text matching~\cite{scan} and propose an automatic evaluation metric for image captioning based on the matching results. Our goal is to capture comprehensive information from input data while also providing an explainable method to assess the quality of an image description.

\begin{figure}
    \centering
    \includegraphics[scale = 0.28]{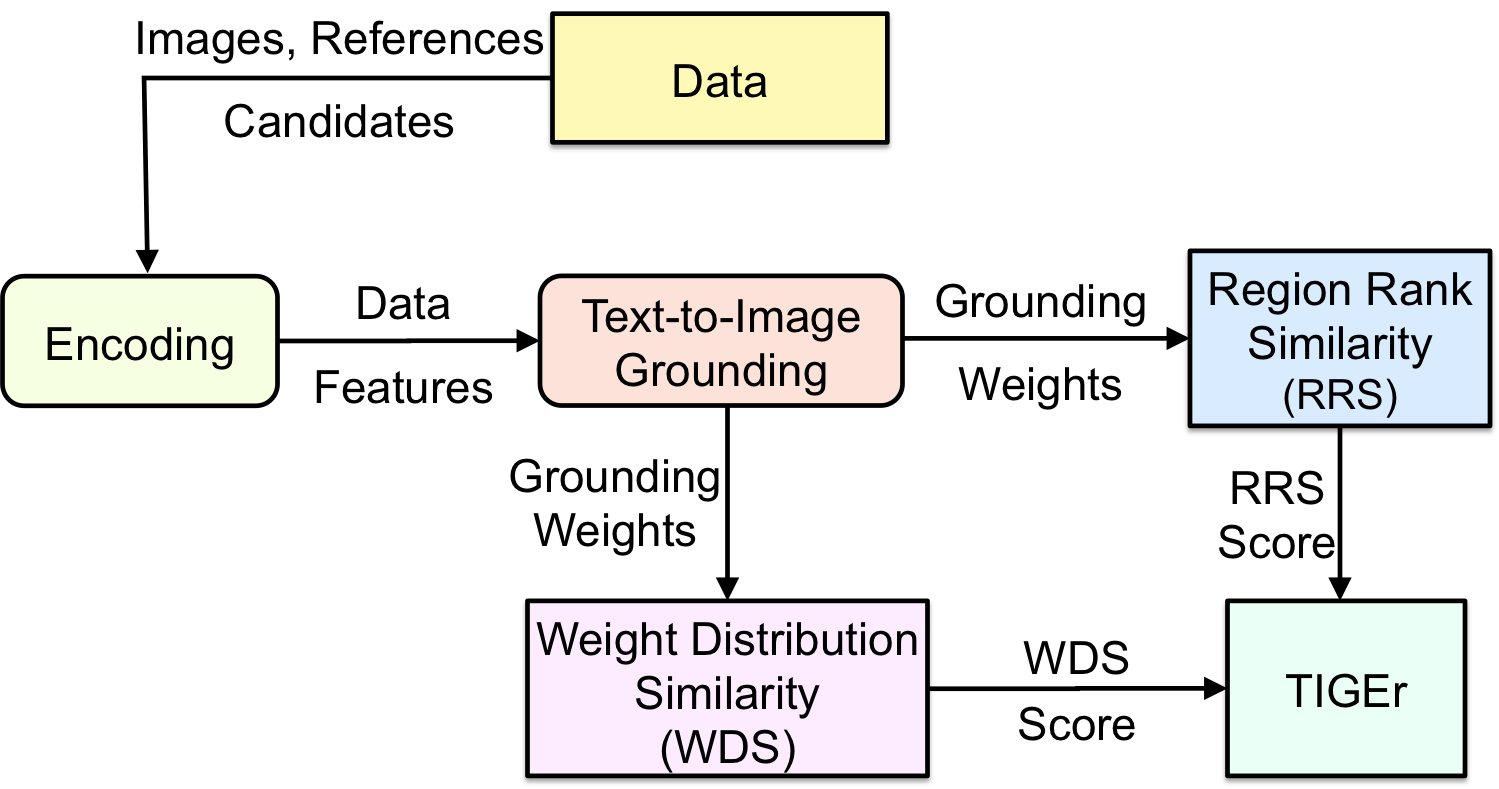}
    \caption{TIGEr framework.
    }
    \label{fig:framework}
    \vspace{-3ex}
\end{figure}

\begin{figure*}[!htb]
    \centering
    \includegraphics[scale = 0.50]{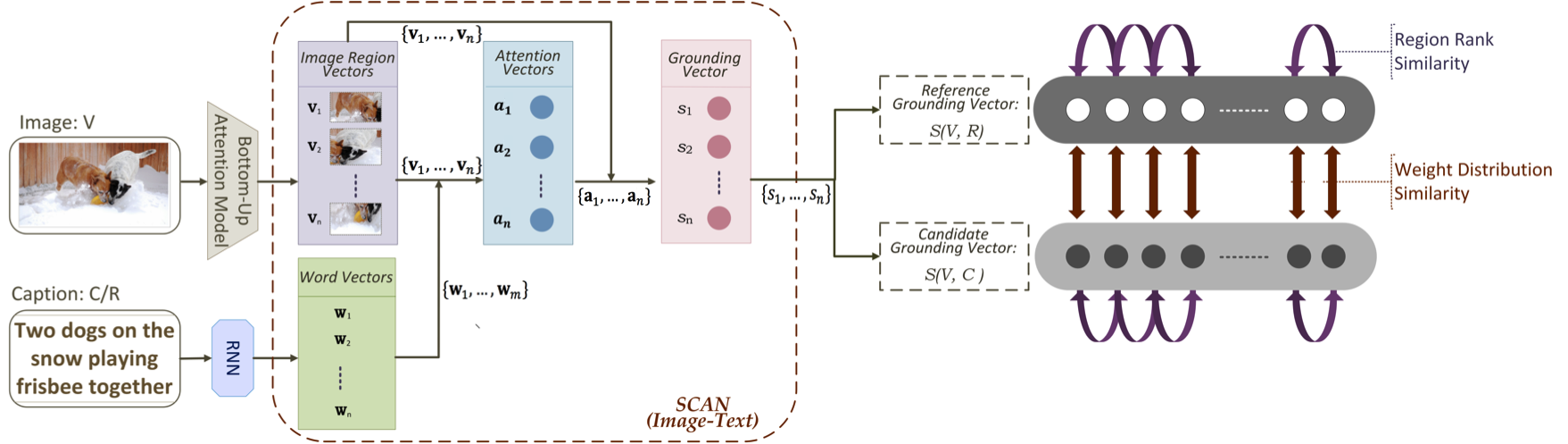}
    \caption{Overview of TIGEr calculation. For each pair of image and caption sentence, the pre-trained SCAN model generates a similarity vector where each dimension denotes the grounding relevance between an image region and the caption sentence in this region context. 
    Given two similarity vectors denoting the grounding outcomes of reference versus candidate captions, we measure: 1) region rank disagreement, and 2) the similarity of two grounding weight distributions.
    %\xin{the font in the figure is still too small.}
    }
    \label{fig:scan}
    \vspace{-2ex}
\end{figure*}

\section{The TIGEr Metric}
%\textcolor{orange}{
The overall framework of TIGEr is shown in Figure~\ref{fig:framework}. 
With the assumption that a good machine-generated caption $C$ should generate a description of an image $V$ like a human would, we compare $C$ against a set of reference captions produced by humans $\mathcal{R}=\{R_1,...,R_k\}$ in two stages.
%}

%\textcolor{orange} {
The first stage is \emph{text-image grounding}, where for each caption-image pair, we compute a vector of \emph{grounding} scores, one for each specific region of the image, indicating how likely the caption is grounded in the region. The grounding vector can also be interpreted as estimating how much attention a human judge or the to-be-evaluated image captioning system pays on each image region in generating the caption. A good system is expected to distribute its attention among different regions of an image similarly to that of human judges. For example, if a human judge wrote a caption solely based on a particular region of an image (e.g., a human face or a bottle as shown in Figure~\ref{fig:demo}) while ignoring the rest of the image, then a good machine-generated caption would be expected to describe only the objects in the same region. 
%}

%\textcolor{orange}{
The second stage is \emph{grounding vector comparison}, where we compare the grounding vector between an image $V$ and the machine-generated caption $C$, denoted by $\mathbf{s}(V,C)$, and that between $V$ and reference captions $\mathcal{R}$, denoted by $\mathbf{s}(V,\mathcal{R})$. The more similar these two vectors are, the higher quality of $C$ is. The similarity is measured by using two metric systems. The first one measures how similarly these image regions are ordered (by their grounding scores) in the two vectors. The second one measures how similarly the attention (indicated by grounding scores) is distributed among different regions of the image in the two vectors. The TIGEr score is the average score of the resulting two similarity scores. 
%}

In the rest of this section, we describe the two stages in detail.

%match reference/candidate captions with a fixed number of regions per image in a semantic space by building a text-to-image grounding module. 
%Second, we design two comparison modules: 1) the score of region rank difference measures the ranking disagreement of image regions according to their relevance to the text in the grounding process between references and candidates. Inspired by the phenomenon that humans prefer to describe the primary parts of an image such as objects, actions and locations rather than all content, this module tends to capture the consensus of relevance priority over image regions between a machine-generated caption and human-written references. 2) The score of weight distribution difference quantifies the similarity between reference and candidate grounding weight distribution. This comparison is motivated by the consideration that even with the same order of relevance across image regions, the matching value per region grounded with candidate versus reference captions can be various. Finally, TIGEr outputs the average score of both comparison results.}

\subsection{Text-Image Grounding}
%In order to have a fine-grained grounding from the words in a caption to its associated image regions, we adopt a state-of-the-art model to ground image data with text data,  namely Stacked Cross Attention Neural Network (SCAN)~\cite{scan}. We pre-train the SCAN model on the 2014 MS-COCO train/val dataset with default settings. The training data contains 121,287 images, of which each is annotated with five text descriptions.
To compute the grounding scores of an image-caption pair, we need to map the image and its paired caption into vector representations in the same vector space. We achieve this by using a Stacked Cross Attention Neural Network (SCAN)~\cite{scan}, which is pre-trained on the 2014 MS-COCO train/val dataset with default settings\footnote{Github link: https://github.com/kuanghuei/SCAN}. The training data contains 121,287 images, of which each is annotated with five text descriptions. The architecture of SCAN is shown in Figure~\ref{fig:scan}.

We first encode caption $C$ or $R$, which is a sequence of $m$ words, into a sequence of $d$-dimensional ($d=300$) word embedding vectors, $\{\mathbf{w}_1,...,\mathbf{w}_m\}$, using a bi-directional recurrent neural network (RNN) models with gated recurrent units (GRU) \cite{rnngru}. 
We then map image $V$ into a sequence of feature vectors in two steps. First, we extract from $V$ a set of $n=36$ region-level 2048-dimensional feature vectors using a pre-trained bottom-up attention model \cite{bottomup}. Then, we use linear projection to convert these vectors into a sequence of $d$-dimensional image region vectors $\{\mathbf{v}_1,...,\mathbf{v}_n\}$.

% \textcolor{orange}{First, we extract 36 region-level image features from a target image \(I\) by using a pre-trained bottom-up attention model \cite{bottomup}, where each 2048-dimensional feature vector represents a region of \(I\).} Meanwhile, text features \(W = \{w_1, v_2,...,w_m\}, w_j \in \mathbb{R}^D\) are encoded from a caption sentence \(T\) by employing a bi-directional gated recurrent units (GRU) \cite{rnngru}, where \(w_j\) denotes a 300-dimensional word vector. \textcolor{orange}{In order to embed image and text features into the same space, we use an linear layer in SCAN model to project image features to the same space of text features, i.e., \(V = \{v_1, v_2,...,v_n\}, v_i \in \mathbb{R}^D\) , where \(v_i\) is a 300-dimensional vector representing the \(i^{th}\) image region (a detected bounding box).} The sentence \(T\) can be either a human-written caption from the reference set \(REF = \{r_1, r_2,...,r_k\}\) or a candidate caption \(c\).

We compute how much caption $C$, represented as $\{\mathbf{w}_1,...,\mathbf{w}_m\}$, is grounded to each image region $\mathbf{v}_i$, $i=1...n$, as follows. First, we compute an attention feature vector for each $\mathbf{v}_i$ as

%\lei{Please note that I changed $\mathbf{a}$ to $\alpha$ in Eq. 2, and removed the transpose symbol in Eq. 1 and 6 for simplicity.}
\vspace{-3.5ex}
\begin{align}
\label{eq3}
\mathbf{a}_i &= \sum_{j=1}^m \alpha_{ij} \mathbf{w}_j \\
\quad \alpha_{ij} &= \frac{\exp(\lambda \text{sim}(\mathbf{v}_i, \mathbf{w}_j))}{\sum_{k = 1}^m \exp(\lambda \text{sim}(\mathbf{v}_i, \mathbf{w}_k))}
\end{align}
where $\lambda$ is a smoothing factor, and $\text{sim}(\mathbf{v}, \mathbf{w})$ is a normalized similarity function defined as
\vspace{-0.25ex}
\begin{equation}\label{eq2}
    \text{sim}(\mathbf{v}_i, \mathbf{w}_j) = \frac{\max (0, \text{score}(\mathbf{v}_i, \mathbf{w}_j))} {\sqrt{\sum_{k = 1}^n \max(0, \text{score}(\mathbf{v}_k, \mathbf{w}_j)) ^2}}
\end{equation}

\begin{equation}\label{eq1}
    \text{score}(\mathbf{v}_i,\mathbf{w}_j) = \frac{\mathbf{v}_i \cdot \mathbf{w}_j}{\parallel \mathbf{v}_i \parallel\parallel \mathbf{w}_j \parallel}
\end{equation}
Then, we get the grounding vector of $C$ and $V$ as
\begin{align}
\label{eq5}
    \mathbf{s}(V, C) = \{s_1,...,s_n\} \\
    s_i:= s(\mathbf{v}_i, C) = \frac{\mathbf{v}_i \cdot \mathbf{a}_i}{\parallel \mathbf{v}_i \parallel\parallel \mathbf{a}_i \parallel}
\end{align}
where the absolute value of $s_i$ indicates how much caption $C$ is grounded to the $i$-th image region or, in other words, to what degree $C$ is generated based on the $i$-th image region.

Given an image $V$, in order to evaluate the quality of a machine-generated caption $C$ based on a set of reference captions $\mathcal{R}=\{R_1,...,R_k\}$, we generate two grounding vectors, $\mathbf{s}(V, C)$ as in Equation~\ref{eq5} and $\mathbf{s}(V, \mathcal{R})$ which is a mean grounding vector over all references in $\mathcal{R}$ as % in Equation~\ref{eq7}.
\begin{equation}\label{eq7}
    \mathbf{s}(V, \mathcal{R}) = \text{avg}([\mathbf{s}(V,R_1),...,\mathbf{s}(V,R_k)]).
\end{equation}

\subsection{Grounding Vector Comparison}
% The comparison of grounding outputs in terms of two similarity vectors \(S(V, REF)\) and \(S(V, c)\) is considered from both relative and absolute perspectives (see Figure~\ref{fig:scan}).
The quality of $C$ is measured by comparing $\mathbf{s}(V,C)$ and $\mathbf{s}(V,\mathcal{R})$ using two metric systems, Region Rank Similarity (RRS) and Weight Distribution Similarity (WDS).

\subsubsection{Region Rank Similarity (RRS)}

RRS is based on Discounted Cumulative Gain (DCG) \cite{ndcg}, which is widely used to measure document ranking quality of web search engines. Using a graded relevance scale of documents in a rank list returned by a search engine, DCG measures the usefulness, or \emph{gain}, of a document based on its position in the ranked list. 

In the image captioning task, we view a caption as a query, an image as a collection of documents, one for each image region, and the grounding scores based on reference captions as human-labeled relevance scores. 
Note that $\mathbf{s}(V,C)$ consists of a set of similarity scores $\{s_1, s_2, ..., s_n\}$ for all image regions. If we view $s(\mathbf{v},C)$ as a relevance score between caption (query) $C$ and image region (document) $\mathbf{v}$, we can sort these image regions by their scores to form a ranked list, similar to common procedure in Information Retrieval of documents from data collections via search engines. 

Then, the quality of the ranked list, or equivalently the quality of $C$, can be measured via DCG, which is the sum of the graded relevance values of all image regions in $V$, discounted based on their positions in the rank list derived from $\mathbf{s}(V,C)$:
\vspace{-1ex}
\begin{equation}\label{eq9}
    \text{DCG}_{\mathbf{s}(V,C)} = \sum_{k=1}^n\frac{rel_{k}}{\log_2 (k + 1)}
\end{equation}
where $rel_k$ is the human-labeled graded relevance value of the image region at position $k$ in the ranked list.

Similarly, we can generate an ideal ranked list from $\mathbf{s}(V,\mathcal{R})$ where all regions are ordered based on their human-labeled graded relevance values. We compute the Ideal DCG (IDCG) as
\vspace{-1ex}
\begin{equation}\label{eq8}
    \text{IDCG}_{\mathbf{s}(V,\mathcal{R})} = \sum_{k=1}^n\frac{rel_{k}}{\log_2 (k + 1)}
\end{equation}
Finally, RRS between $\mathbf{s}(V,C)$ and $\mathbf{s}(V,\mathcal{R})$ is defined as Normalized DCG (NDCG) as
\begin{equation}\label{eq10}
\resizebox{0.85\hsize}{!}{$\text{RRS}_{(V,C,\mathcal{R})} := \text{NDCG}_{(V,C,\mathcal{R})} = \frac{\text{DCG}_{\mathbf{s}(V,C)}}{\text{IDCG}_{\mathbf{s}(V,\mathcal{R})}}$}
\end{equation}

The assumption made in using RRS is that $C$ is generated based mainly on a few highly important image regions, rather than the whole image. A high-quality $C$ should be generated based on a similar set or the same set of highly important image regions based on which human judges write reference captions. 
One limitation of RRS is that it is not suitable to measure the quality of captions that are generated based on many equally important image regions. In these cases, we assume that humans produce captions by distributing their attention more or less evenly across all image regions rather than focusing on a small number of highly important regions. The remedy to this limitation is the new metric we describe next.

\subsubsection{Weight Distribution Similarity (WDS)}
WDS measures how similarly a system and human judges distribute their attention among different regions of an image when generating captions.
Let $P$ and $Q$ be the attention distributions derived from the grounding scores in $\mathbf{s}(V,\mathcal{R})$ and $\mathbf{s}(V,C)$, respectively. We measure the distance between the two distributions via KL Divergence \cite{kld1} as
\vspace{-1ex}
\begin{align}
\label{eq11}
     D_{KL}(P||Q) &= \sum_{k=1}^n P(k)\log\frac{P(k)}{Q(k)}
\end{align}
where $P(k) = \text{softmax}(\mathbf{s}(\mathbf{v}_k,\mathcal{R}))$ and $Q(k) = \text{softmax}(\mathbf{s}(\mathbf{v}_k,C))$. 

In addition, we also find it useful to capture the difference in caption-image relevance between $(V, C)$ and $(V,\mathcal{R})$. The relevance value of a caption-image pair can be approximated by using the module of each grounding vector. We use $D_{rel}$ to denote the value difference term, defined as 
%The relevance value of a caption-image pair can be approximated using the sum of grounding scores over all image regions stored in its grounding vector. We use $D_{rel}$ to denote the value difference term, defined as 
\begin{align}
\label{rel-diff}
 %D_{rel}((V,\mathcal{R})||(V, C)) &= \log \frac{\sum_{k=1}^n s(\mathbf{v}_k,\mathcal{R})}{\sum_{k=1}^n s(\mathbf{v}_k,C)}
 D_{rel}((V,\mathcal{R})||(V, C)) &= \log \frac{\|\mathbf{s}(V,\mathcal{R})\|}{\|\mathbf{s}(V,C)\|}
\end{align}

Finally, letting $D(\mathcal{R}||C)=D_{KL}(P||Q) + D_{rel}((V,\mathcal{R})||(V, C))$,
we get WDS between two grounding vectors using a sigmoid function as
% \footnote{The detailed measurement of WDS is provided in appendix}
\begin{equation}
    \text{WDS}_{(V,C,\mathcal{R})} :=1-  \frac{\exp{(\tau {D(\mathcal{R}||C)})}}{\exp{(\tau {D(\mathcal{R}||C)})} + 1}
\end{equation}
where $\tau$ is the to-be-tuned temperature. 

\subsubsection{TIGEr Score}
The TIGEr score is defined as the average value\footnote{We selected arithmetic mean by empirically observing the value variance between RRS and WDS in the [0, 1] interval.}of RRS and WDS:
\begin{equation}\label{eq12}
    \resizebox{0.85\hsize}{!}{$\text{TIGEr}_{(V,C,\mathcal{R})} := \frac{\text{RRS}_{(V,C,\mathcal{R})} + \text{WDS}_{(V,C,\mathcal{R})}}{2}$}
\end{equation}
The score is a real value of $[0,1]$. A higher TIGEr score indicates a better caption as it matches better with the human generated captions for the same image. 

\section{Experiments}
%\subsection{Experimental Setup}
\subsection{Dataset}
\paragraph{Composite Dataset}
The multisource dataset \cite{composite} we used contains testing captions for 2007 MS-COCO images, 997 Flickr 8k pictures, and 991 Flickr 30k images. In total, there are 11,985 candidates graded by annotators on the description relevance from 1 (not relevant) to 5 (very relevant). Each image has three candidate captions, where one is extracted from human-written references, and the other two are generated by recently proposed captioning models \cite{composite_data1, composite_data2}.

\paragraph{Flickr 8K}
The Flickr 8K dataset was collected by \citet{flickr8k}, and contains 8092 images. Each picture is associated with five reference captions written by humans. This dataset also includes 5822 testing captions for 1000 images. Unlike the aforementioned two datasets, where testing captions are directly generated based on an image, candidates in Flickr 8K were selected by an image retrieval system from a reference caption pool. For grading, native speakers were hired to give a score from 1 (not related to the image content) to 4 (very related). Each caption was graded by three human evaluators, and the inter-coder agreement was around 0.73. Because 158 candidates are actual references of target images, we excluded these for further analysis.

\paragraph{Pascal-50S}
The PASCAL-50S dataset \cite{cider} contains 1000 images from the UIUC PASCAL Sentence dataset, of which 950 images are associated with 50 human-written captions per image as references, and the remainder of the images has 120 references for each picture. This dataset also contains 4000 candidate caption pairs with human evaluation, where each annotator was asked to select one candidate per pair that is closer to the given reference description. Candidate pairs are grouped by four types: 1) human-human correct (HC) contains two human-written captions for the target image, 2) human-human incorrect (HI) includes two captions written by human but describing different images, 3) the group of human-machine (HM) is contains a human-written and a machine-generated caption, and 4) machine-machine (MM) includes two matching-generated captions focusing on the same image.

\subsection{Compared Metrics}
Given our emphasis on metric interpretability and efficiency, we selected six rule-based metrics that have been widely used to evaluate image captions for comparison. The metrics are BLEU-1, BLEU-4, ROUGE-L, METEOR, CIDEr, and SPICE. We use MS COCO evaluation tool\footnote{https://github.com/tylin/coco-caption} to implement all metrics. Before testing, the input texts were lowercased and tokenized by using the ptbtokenizer.py script from the same tool package.

\subsection{Evaluation}\label{eval}
Our examination of metric performances mainly focuses on the caption-level correlation with human judgments. Following prior studies \cite{spice, evlauation1}, we use Kendall's tau (\(\tau\)) and Spearman's rho ($\rho$) rank correlation to evaluate pairwise scores between metrics and human decisions in the Composite and Flickr 8K datasets. 
\begin{table}[!hbt]
% \small
% \setlength{\tabcolsep}{4pt}
\centering
\resizebox{0.75\columnwidth}{!}{%
\begin{tabular}{ l c  c  c  c }
    \toprule
    & \multicolumn{2}{c}{Composite} & \multicolumn{2}{c}{Flickr8k} \tabularnewline
	 \cmidrule(lr){2-3}\cmidrule(lr){4-5}
	 &  $\tau$ & $\rho$ & $\tau$ & $\rho$ \tabularnewline
	 \midrule
	BLEU-1 & 0.280 & 0.353 & 0.323 & 0.404\tabularnewline
	BLEU-4 & 0.205 & 0.352 & 0.138 & 0.387\tabularnewline
	ROUGE-L & 0.307 & 0.383 & 0.323 & 0.404\tabularnewline
	METEOR & 0.379 & 0.469 & 0.418 & 0.519 \tabularnewline
	CIDEr & 0.378 & 0.472 & 0.439 & 0.542 \tabularnewline
	SPICE & 0.419 & 0.514 & 0.449  & 0.596  \tabularnewline
	\midrule
	\textbf{Ours} & \tabularnewline
	RRS & 0.388 & 0.479 & 0.418 & 0.521\tabularnewline
% 	Absolute Diff. & 0.441 & 0.441 & 0.409 & 0.409 \tabularnewline
	WDS & 0.433 & 0.526 & 0.464 & 0.572\tabularnewline
% 	TIGEr & 0.442 & 0.442 & 0.412  & 0.412  \tabularnewline 
	TIGEr & \textbf{0.454} & \textbf{0.553} &\textbf{0.493} & \textbf{0.606} \tabularnewline 
\bottomrule
\end{tabular}
}
 \caption{Caption-level correlation between metrics and human grading scores in Composite and Flickr 8K dataset by using Kendall tau and Spearman rho. All p-values \(<\) 0.01.
 }\label{tab:kendall}
 \vspace{-2ex}
 \end{table}
 
\begin{figure}[!hbt]
\vspace{-1ex}
   \centering
     \includegraphics[width=0.48\textwidth, height = 3.5 cm]{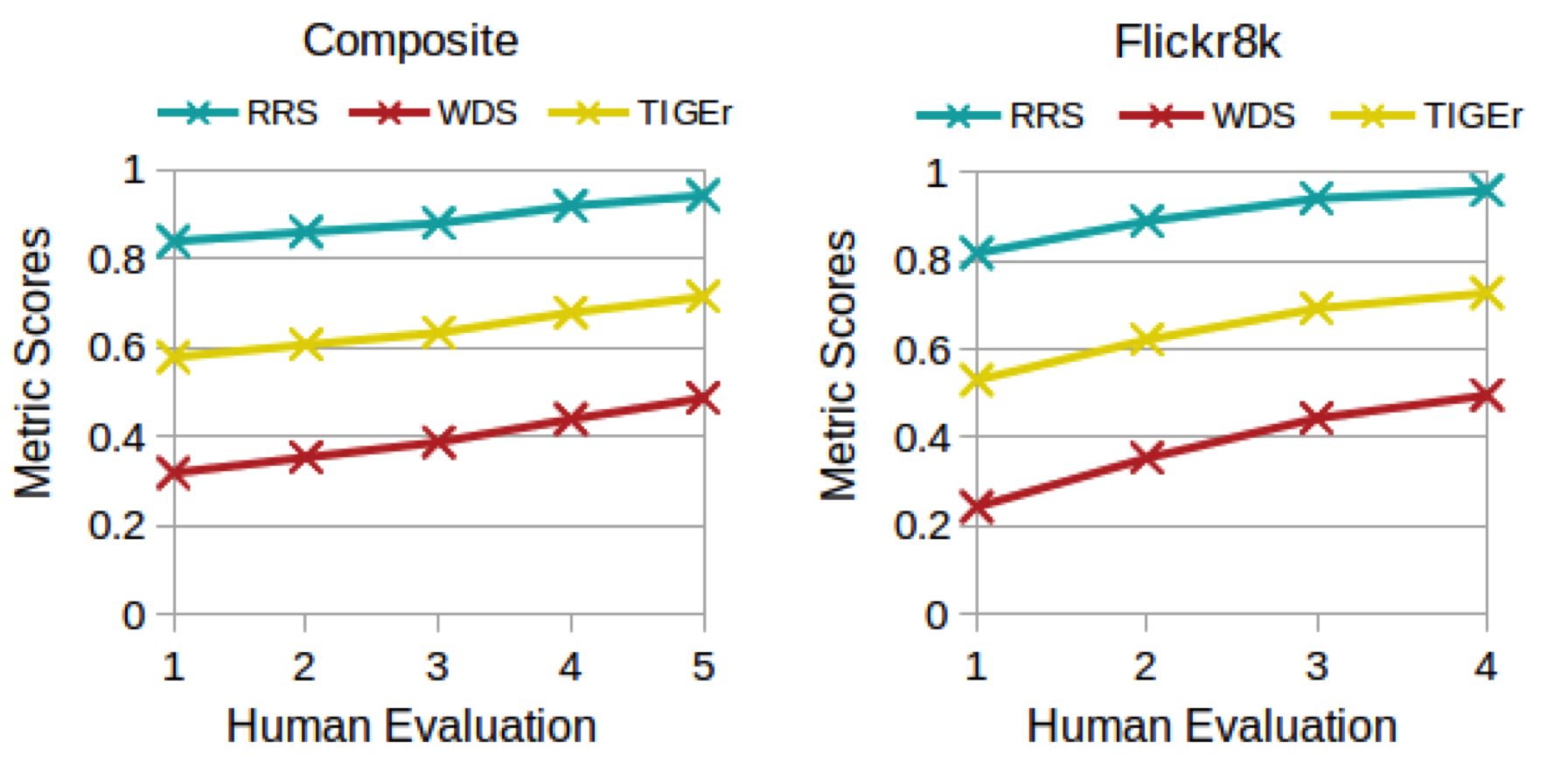}
     \caption{Average metric score based on human score group.}
   \vspace{-1.5ex}
   \label{fig:mscore}
\end{figure}  
Since human annotation of the Pascal-50S data is a pairwise ranking instead of scoring, we kept consistency with the prior work \cite{cider, spice} that evaluates metrics by accuracy. Different from \citet{spice} who considered equally-scored pairs as correct cases, our definition of a correct case is that the metric should assign a higher score to a candidate that was preferred by human annotators.

\subsection{Result on Composite \& Flickr 8K}
\paragraph{Metric Performance} 
Table~\ref{tab:kendall} displays the correlation between metrics and human judgments in terms of $\tau$ and $\rho$.
%\(\tau_C\) and \(rev\_\tau_C\) (as described in Sec.~\ref{eval}). 
Based on both correlation coefficients, we achieved a noticeable improvement in the assessment of caption quality on Composite and Flick8K compared to the previous best results produced with existing metrics \cite{spice, evlauation1}. Regarding the isolated impact of two similarity measures, we observe that WDS contributes more to caption evaluation than RRS. This finding indicates that the micro-level comparison of grounding similarity distribution is more sensitive to human judges than the macro-level contrast of image region rank by grounding scores.

\paragraph{TIGEr Score Analysis} To understand the evaluation results of TIGEr, we further analyzed our metric scores based on the group of human scoring on caption quality. As shown in Figure~\ref{fig:mscore}, our metric scores are increasing according to the growth of human scores in both datasets. Overall, the growth rate of RRS is similar with the WDS score per dataset. Interestingly, the value of the metric scores increase more slowly in high-scored caption groups than in low-scored groups (Flickr8k), which suggests that the difference in caption relevance between adjacent high-scored groups (e.g., 3 vs. 4) is smaller than the descriptions in nearby low-scored groups. A more detailed error analysis and qualitative analysis is addressed in the Appendix.

%tent.paslieveca Be exl-50 accuracy
\begin{figure*}[t]
\centering
   \begin{minipage}{0.48\textwidth}
     \centering
     \includegraphics[scale = 0.45]{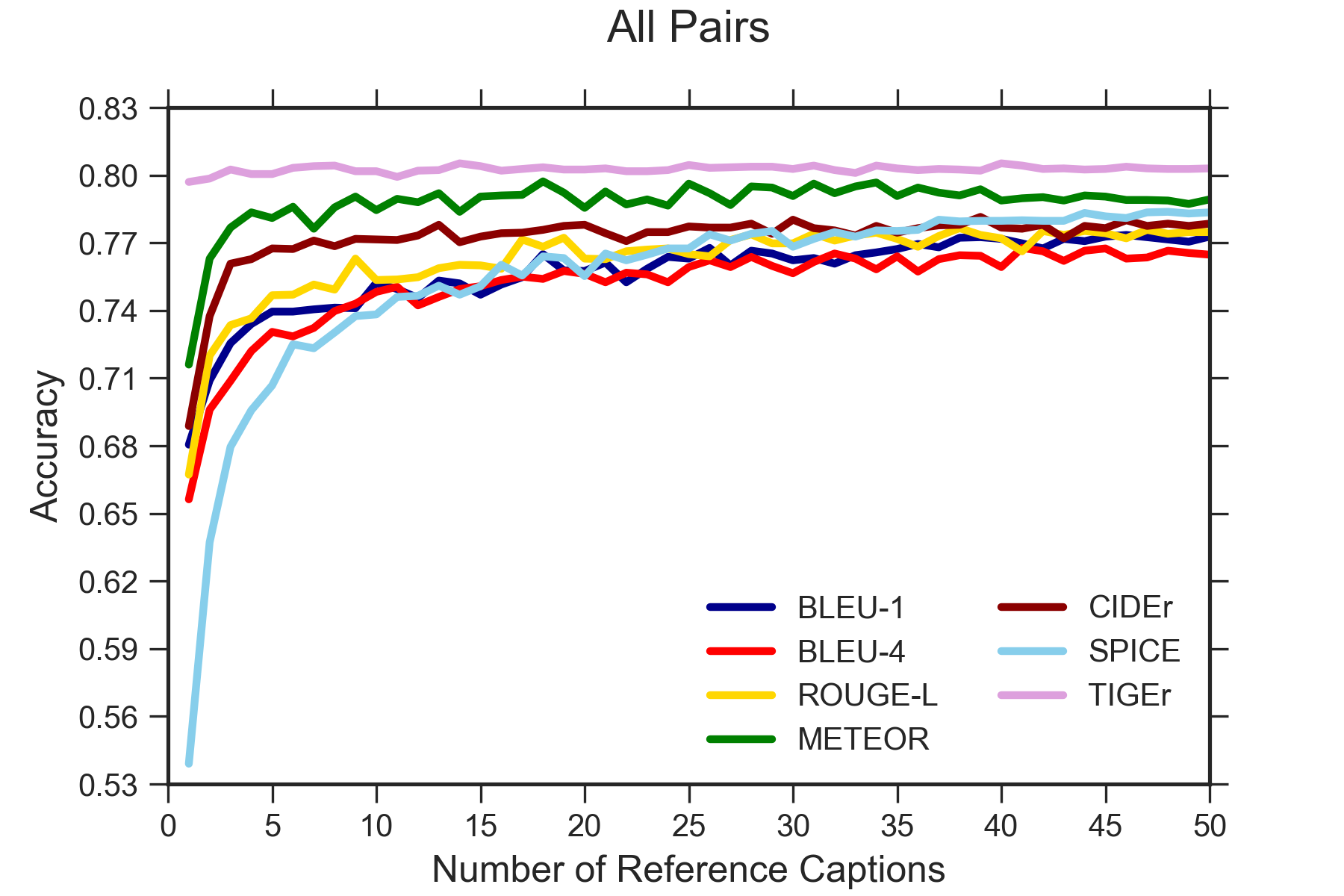}
    %\caption{Interpolation for Data 1}\label{Fig:Data1}
   \end{minipage}\hfill
   \begin{minipage}{0.48\textwidth}
     \centering
     \includegraphics[scale = 0.45]{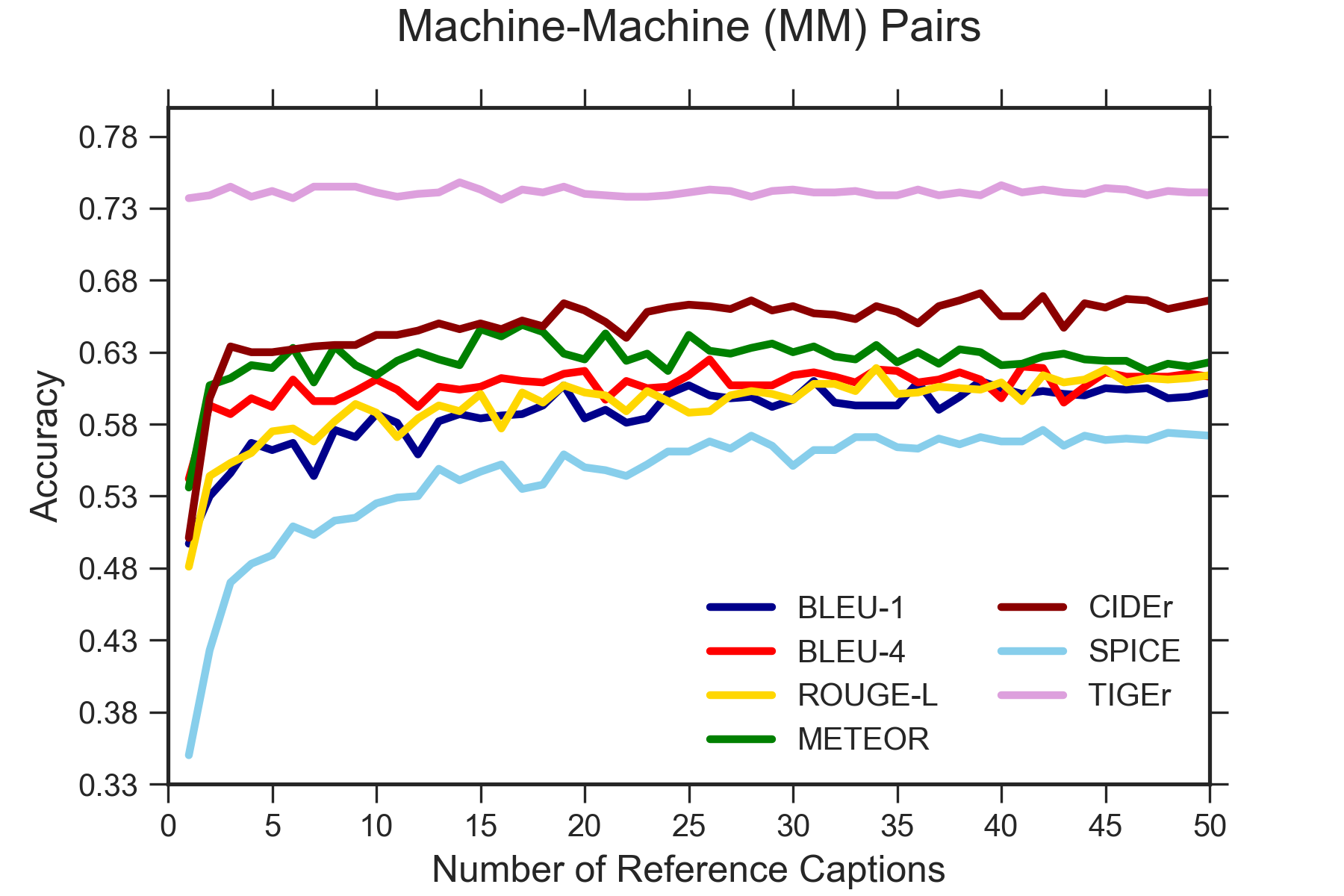}
     %\caption{Interpolation for Data 2}\label{Fig:Data2}
   \end{minipage}
   %\vspace{-1.5ex}
   \caption{Pairwise comparison accuracy (y-axis) of metrics at matching human judgments with 1-50 reference captions (x-axis). TIGEr (solid line in light-purple) is the best performing metric in both all pair group and machine-machine pair group. The number of reference captions is obviously beneficial to all existing metrics except TIGEr, which is less dependent on excessive references and can therefore greatly reduce human annotation costs.
   }
   \label{Fig:Pascal_MultipleReferences}
\end{figure*}

\begin{figure*}[t]
    \centering
    \includegraphics[width = 0.88\textwidth, height = 3 cm]{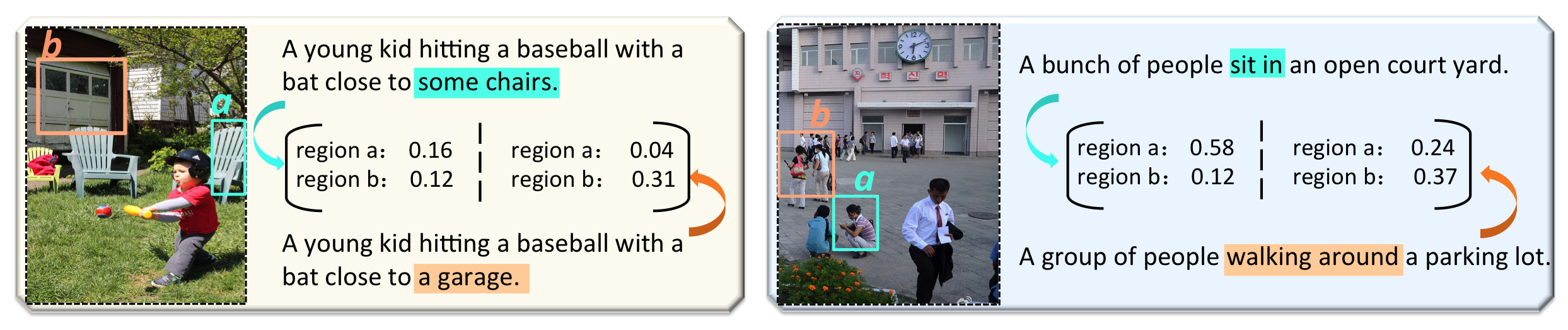}
    \vspace{-1ex}
    \caption{Visualization of text-to-image grounding.}
    \vspace{-1.5ex}
    \label{fig:case}
\end{figure*}

\subsection{Result on Pascal-50S}
\paragraph{Fixed Reference Number} 
Table~\ref{tab:pascal_5} reports the evaluation accuracy of metrics on PASCAL-50S with five reference captions per image. Our metric achieves higher accuracy in most pair groups except for HM and HC. Given all instances, TIGEr improves the closeness to human judgment by 2.72\% compared to the best prior metric (i.e., METEOR). Among the four considered candidate groups, identifying irrelevant human-written captions in HI is relatively easy for all metrics, and TIGEr achieves the highest accuracy (99.80\%). In contrast, judging the quality of two correct human-annotated captions in HC is difficult with a lower accuracy per metric compared to other testing groups. For this pair group, TIGEr (56.00\%) shows a comparable performance with the best alternative metric (METEOR: 56.70\%). More importantly, TIGEr reaches a noteworthy improvement in judging machine-generated caption pairs (MM) with an increasing in terms of accuracy by about 10.00\%  compared to the best prior metric (CIDEr: 64.50\%).

\paragraph{Changing Reference Number}
In order to explore the impact of reference captions on metric performance, we changed the number of references from 1 to 50. As shown in Figure~\ref{Fig:Pascal_MultipleReferences}, TIGEr outperforms prior metrics in All and MM candidate pairs by: 1) achieving higher accuracy, especially for small numbers of references; and 2) obtaining more stable performance results across varied reference sizes. Our findings suggest that TIGEr has low reference dependency. Compared with prior work~\cite{cider}, the slight differences in results might be caused by the random choices for reference subsets.
\begin{table}[t!]
%\vspace{-2ex}
\setlength{\tabcolsep}{5pt}
\centering
% \scalebox{0.85}{
\resizebox{0.85\columnwidth}{!}{%
\begin{tabular}{  l  c  c  c  c  c }
    \toprule
	\textbf{} & \textbf{HC} & \textbf{HI} & \textbf{HM} & \textbf{MM} & \textbf{All} \\ 
	\midrule
	BLEU-1 & 51.20 & 95.70 & 91.20 & 58.20 & 74.08\\
	BLEU-4 & 53.00 & 92.40 & 86.70 & 59.40 & 72.88\\ 
	ROUGE-L & 51.50 & 94.50 & 92.50 & 57.70 & 74.05\\
	METEOR & \textbf{56.70} & 97.60 & \textbf{94.20} & 63.40 & 77.98\\
	CIDEr & 53.00 & 98.00 & 91.50 & 64.50 & 76.75\\
	SPICE & 52.60 & 93.90 & 83.60 & 48.10 & 69.55\\
	\midrule
	TIGEr (ours) & 56.00 & \textbf{99.80} & 92.80 & \textbf{74.20} & \textbf{80.70}\\ 
    \bottomrule
\end{tabular}
}
\caption{Accuracy of metrics at matching human judgments on PASCAL-50S with 5 reference captions. The highest accuracy per pair type is shown in bold font. Column titles are explained in Section 4.1.}
\label{tab:pascal_5}
\vspace{-1ex}
\end{table}

\subsection{Text Component Sensitivity}
\begin{figure}[t!]
     \centering
     \includegraphics[width=0.45\textwidth, height = 4 cm]{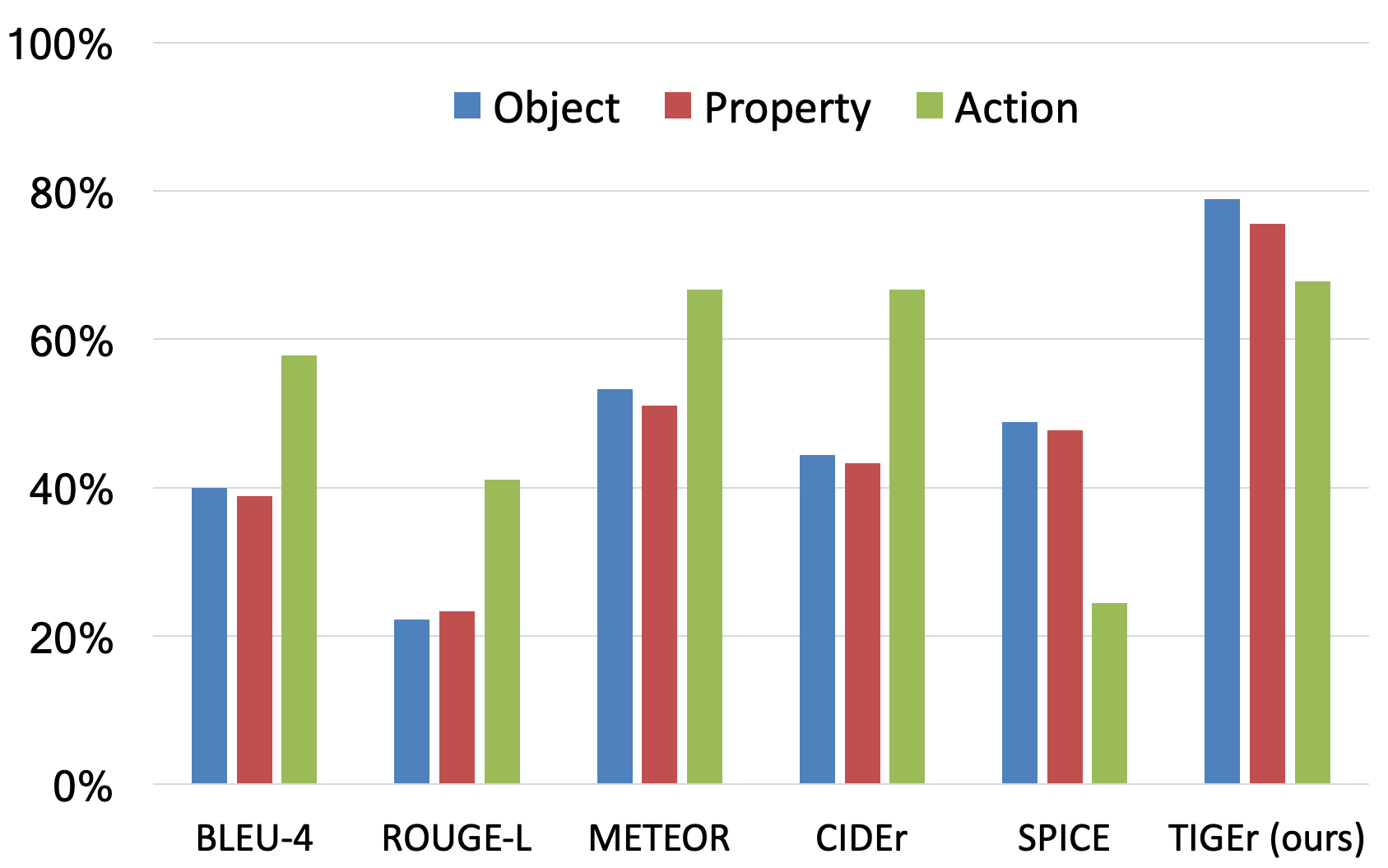}
     \caption{Metric accuracy at three text component levels.
     }
     \label{micro_result}
%\vspace{-3ex}
\end{figure}

We also explore the metrics' capability to identify specific text component errors (i.e., object, action, and property). We randomly sampled a small set of images from Composite and Pascal-50S. Given an image, we pick a reference caption and generate two candidates by replacing a text component. For example, we replaced the action \textit{``walk"} from a reference \textit{``People walk in a city square"} with a similar action \textit{``stroll"} and a different action\textit{``are running"} as a candidate pair. We then calculated the accuracy of pairwise ranking per metric for each component. As Figure~\ref{micro_result} shows, TIGEr is sensitive to recognizing object-level changes while comparatively weak in detecting action differences. This implies that text-to-image grounding is more difficult at the action-level than the object-level. Similarly, SPICE also has a lower capability to identify action inconsistencies, which shows the limitation of scene graphs to capture this feature. In contrast, the n-gram-based metrics prefer to identify movement changes. To further study the sensitivity of TIGEr to property-level (i.e., object attributes) differences, we manually grouped the compared pairs based on their associated attribute category such as color. Our results show that states (e.g., sad, happy), counting (e.g., one, two), and color (e.g., red, green) are top attribute groups in the sampled data, where TIGEr can better identify the differences of the counting description (acc. 84.21\%) compared to the expression of state (69.70\%) and color (69.23\%).

%\vspace{-1ex}
\subsection{Grounding Analysis}
To analyze the process of text-to-image grounding behind TIGEr in depth, we visualize a set of matching cases of image-caption pairs. Figure~\ref{fig:case} provides two illustrative examples. Interestingly, both captions per image are correct, but focus on different image parts. For instance, in the left image, caption \textit{1} mentions ``chairs", which matches region \textit{a}, while caption \textit{2} relates to region \textit{b} by addressing ``garage". According to our observation, the image region typically has a higher grounding weight with the corresponding caption than the other unrelated region. Our observations suggest that since captions are typically short, they may not cover the information of all regions of an image and hence taking image content into account for captioning evaluation is necessary.

\section{Conclusion and Future Work}
We have presented a novel evaluation metric called TIGEr for image captioning by utilizing text-to-image grounding result based on a pre-trained model. Unlike traditional metrics that are solely based on text matching between reference captions and machine-generated captions, TIGEr also takes the matching between image contents and captions into account , and the similarity between captions generated by human judges and automated systems, respectively. The presented experiments with three benchmark datasets have shown that TIGEr outperforms existing popular metrics, and has a higher and more stable correlation with human judgments. In addition, TIGEr is a fine-grained metric in that it identifies description errors at the object level. 

Though TIGEr was build upon SCAN, our metric is not tied to this grounding model. The improvements with grounding models focusing on the latent correspondence between object-level image regions and descriptions will allow TIGEr to be further improved in the future. Since the pre-trained data mainly contains photo-based images, TIGEr primarily focuses on the caption evaluation in this image domain. For other domains, we can retrain the SCAN model using the ground-truth image-caption pairs, where the process is similar to applying a learning-based metric. In our future work, we plan to extend the metric to other visual-text generation tasks such as storytelling.

\section*{Acknowledgments}
We appreciate anonymous reviewers for their constructive comments and insightful suggestions. This work was mostly done when Ming Jiang was interning at Microsoft Research.

\bibliography{tiger}
\bibliographystyle{acl_natbib}

\newpage
\appendix
\begin{figure*}[t]
\centering
   \begin{minipage}{0.48\textwidth}
     \centering
     \includegraphics[scale = 0.45]{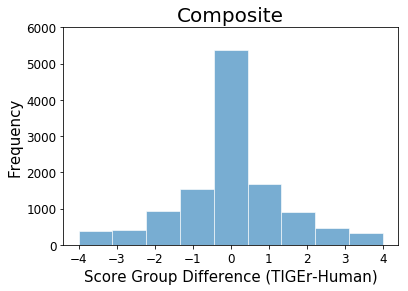}
    %\caption{Interpolation for Data 1}\label{Fig:Data1}
   \end{minipage}\hfill
   \begin{minipage}{0.48\textwidth}
     \centering
     \includegraphics[scale = 0.45]{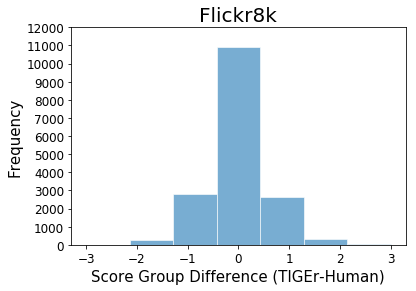}
     %\caption{Interpolation for Data 2}\label{Fig:Data2}
   \end{minipage}
   \vspace{-1.5ex}
   \caption{Distribution of group score differences between TIGEr and human evaluation}
   \label{fig:errdist}
\end{figure*}

\begin{table*}[b]
    \centering
    \begin{tabular}{c}
         \raisebox{-\totalheight}{\includegraphics[width=1.0\textwidth]{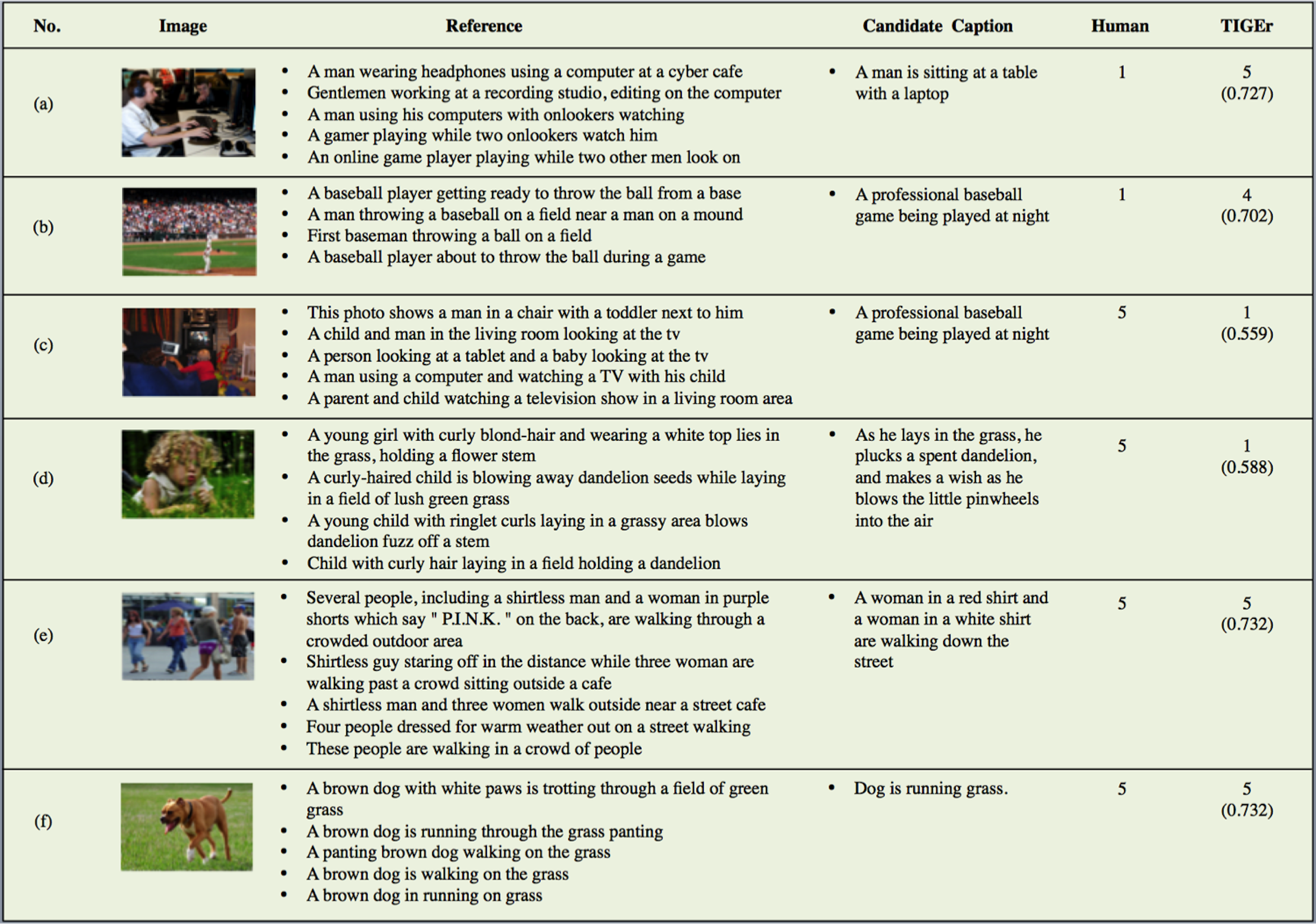}}
    \end{tabular}
    \caption{Examples of scores given by TIGEr. In the column of TIGEr, we display both score group and the actual grading value. The score group is assigned to be comparable with human scores, where we map our metric score into the n-point scale like human evaluation. The detailed mapping process is shown in Section A.1.}
    \label{tab:example}
\end{table*}
\section{Appendices}
\subsection{Error Analysis}
To have a better understanding of inconsistent score ranks between TIGEr and human evaluation assigned to candidates in the Composite and Flick8k datasets, we further conducted an error analysis. Considering the score of TIGEr is  continuous [0, 1], while human scoring is done on an n-point scale, we map our metric score onto the n-point scale by sorting instances according to TIGEr results, and assigning a score group per instance based on the distribution of human scores. Figure~\ref{fig:errdist} shows the distribution of score group differences between TIGEr and human evaluation for both datasets. We observe that true positives (i.e., the score group of TIGEr is equal to the human score for a given caption) achieve higher frequency than each error group. On the other hand, inconsistent cases with the small differences of assigned scores (e.g., TIGEr assigns 3, while human assigns 4 to a testing caption) get a higher frequency than the notable difference between two metric scores, especially for the test cases from the Flickr8k dataset. Our finding suggests that captions with subtle quality difference are more difficult to be discriminated than instances with significant gaps. 

\subsection{Qualitative Analysis}
To illustrate some characteristics of TIGEr more in depth, we show human and TIGEr scores for a set of caption examples in Table~\ref{tab:example}. In order to make the result of two metrics comparable, both the score group and actual grading value of TIGEr are displayed. Note that we provide an equal number of examples for each type of comparison result (i.e., either human or TIGEr score is higher than the other one \& both scores are equal) in Table~\ref{tab:example}. 

We find that TIGEr is able to measure a caption quality by considering the semantic information of image contents. For example, human-written references in case \textit{(e)} primarily provide an overall description of all people shown in the image, while the candidate caption specifically describes two walking women in the image. Despite such difference at the text-level, TIGEr assigns a high score to this sentence - just like human evaluators. Case \textit{(b)} also supports this finding to some extent. Unlike the references that specially described a  baseball payer in the game, the candidate caption in this example provides a general description that is matching with the image content (e.g., ``baseball game'', ``at night''). This observation may help to explain why TIGEr gave this sentence a high score.

Besides, we find two challenges for caption evaluation based on TIGEr. First, though our metric improved evaluation performance by considering semantic information from both images and references, objects with closed semantic meaning such as "laptop" and "computer" in example \textit{(a)} is limited to differentiate. Second, human interpretation inspired by the image is hard to be judged by an automatic evaluation metric. For example, in the case \textit{(d)}, ``\textit{makes a wish} as he blows the little pinwheels into the air'' is a reasonable imagination of an human annotator, which cannot be explicitly observed from the image. 

\end{document}